\documentclass[11pt]{article}

\usepackage[table]{xcolor}
\usepackage{multirow}
\usepackage{booktabs}
\usepackage{makecell}
\usepackage{caption}
\usepackage{graphicx}
\usepackage{array}
\usepackage{threeparttable}
\usepackage{adjustbox}
\usepackage{colortbl}
\usepackage{siunitx}
\usepackage{ulem}
\usepackage{amsmath}
\usepackage{amssymb}
\usepackage{amsthm}
\usepackage{bm} 
\usepackage[utf8]{inputenc}
\usepackage{geometry}
\usepackage[ruled,vlined]{algorithm2e}
\newtheorem{theorem}{Theorem}

\usepackage{subcaption}
\usepackage{graphicx}

\usepackage[final]{acl}
\usepackage{times}
\usepackage{latexsym}

\usepackage[T1]{fontenc}

\usepackage[utf8]{inputenc}

\usepackage{microtype}

\usepackage{inconsolata}

\usepackage{graphicx}
\usepackage{tcolorbox}

\usepackage{colortbl}
\definecolor{forward}{RGB}{84, 130, 53}
\definecolor{inverse}{RGB}{47, 85, 151}
\definecolor{resist}{RGB}{128, 0, 128}
\definecolor{rebound}{RGB}{133, 19, 33}

\definecolor{def}{RGB}{119, 228, 200}
\definecolor{thm}{RGB}{69, 53, 193}
\newtcolorbox{thmbox}[1][]{colback=thm!5!white,colframe=thm!60!black,boxsep=-4pt,grow to left by=4pt,left=10pt,grow to right by=4pt,right=10pt,top=10pt,bottom=10pt,#1}
\newtcolorbox{defbox}[1][]{colback=def!5!white,colframe=def!60!black,boxsep=-4pt,grow to left by=4pt,left=10pt,grow to right by=4pt,right=10pt,top=10pt,bottom=10pt,#1}

\title{
    Agent-Dice: Disentangling Knowledge Updates via Geometric Consensus for Agent Continual Learning
}

\author{Zheng Wu\textsuperscript{1}\thanks{This work was done during Zheng Wu’s internship at OPPO Research Institute.}\quad  Xingyu Lou\textsuperscript{2}\quad Xinbei Ma\textsuperscript{1}\quad  Yansi Li\textsuperscript{1}\quad  Weiwen Liu\textsuperscript{1}\quad Weinan Zhang\textsuperscript{1}\quad \\ \textbf{Jun Wang\textsuperscript{2}}\thanks{Corresponding authors. This work was supported by National Natural Science Foundation of China (62406188), and Natural Science Foundation of Shanghai (24ZR1440300).}\quad \textbf{Zhuosheng Zhang\textsuperscript{1}}\textsuperscript{\textdagger}\\
\textsuperscript{1}School of Computer Science, Shanghai Jiao Tong University \quad \textsuperscript{2}OPPO Research Institute \\
\texttt{\{wzh815918208, sjtumaxb, yansi\_li, wwliu, wnzhang, zhangzs\}@sjtu.edu.cn}
\\
\texttt{louxingyu@oppo.com}\quad
\texttt{junwang.lu@gmail.com}
}

\begin{document}
\maketitle
\begin{abstract}
Large Language Model (LLM)-based agents significantly extend the utility of LLMs by interacting with dynamic environments. 
However, enabling agents to continually learn new tasks without catastrophic forgetting remains a critical challenge, known as the stability–plasticity dilemma.
In this work, we argue that this dilemma fundamentally arises from the failure to explicitly distinguish between common knowledge shared across tasks and conflicting knowledge introduced by task-specific interference. 
To address this, we propose Agent-Dice, a parameter fusion framework based on directional consensus evaluation.
Concretely, Agent-Dice disentangles knowledge updates through a two-stage process: geometric consensus filtering to prune conflicting gradients, and curvature-based importance weighting to amplify shared semantics.
We provide a rigorous theoretical analysis that establishes the validity of the proposed fusion scheme and offers insight into the origins of the stability–plasticity dilemma. 
Extensive experiments on GUI agents and tool-use agent domains demonstrate that Agent-Dice exhibits outstanding continual learning performance with minimal computational overhead and parameter updates.
The codes are available at \url{https://github.com/Wuzheng02/Agent-Dice}.
\end{abstract}

\section{Introduction}

Recent advances in Large Language Models (LLMs) have spurred a paradigm shift in artificial intelligence, empowering agents with robust capabilities in reasoning~\cite{jiang2025developing,plaat2024reasoning}, planning~\cite{wei2025plangenllms,huang2024understanding}, and decision-making~\cite{sun2025llm,huang2025foundation}. 
These agents expand the boundaries of the capabilities of LLMs' by deploying them in dynamic real-world~\cite{chen2026trace,jiang2026principle,jiang2026rmsagen,shi2026codehacker}, specifically by operating graphical user interfaces (GUIs)~\cite{zhang2024large,li2026what} or utilizing tools~\cite{li2025review}.

\begin{figure}
    \centering
    \includegraphics[width=\linewidth]{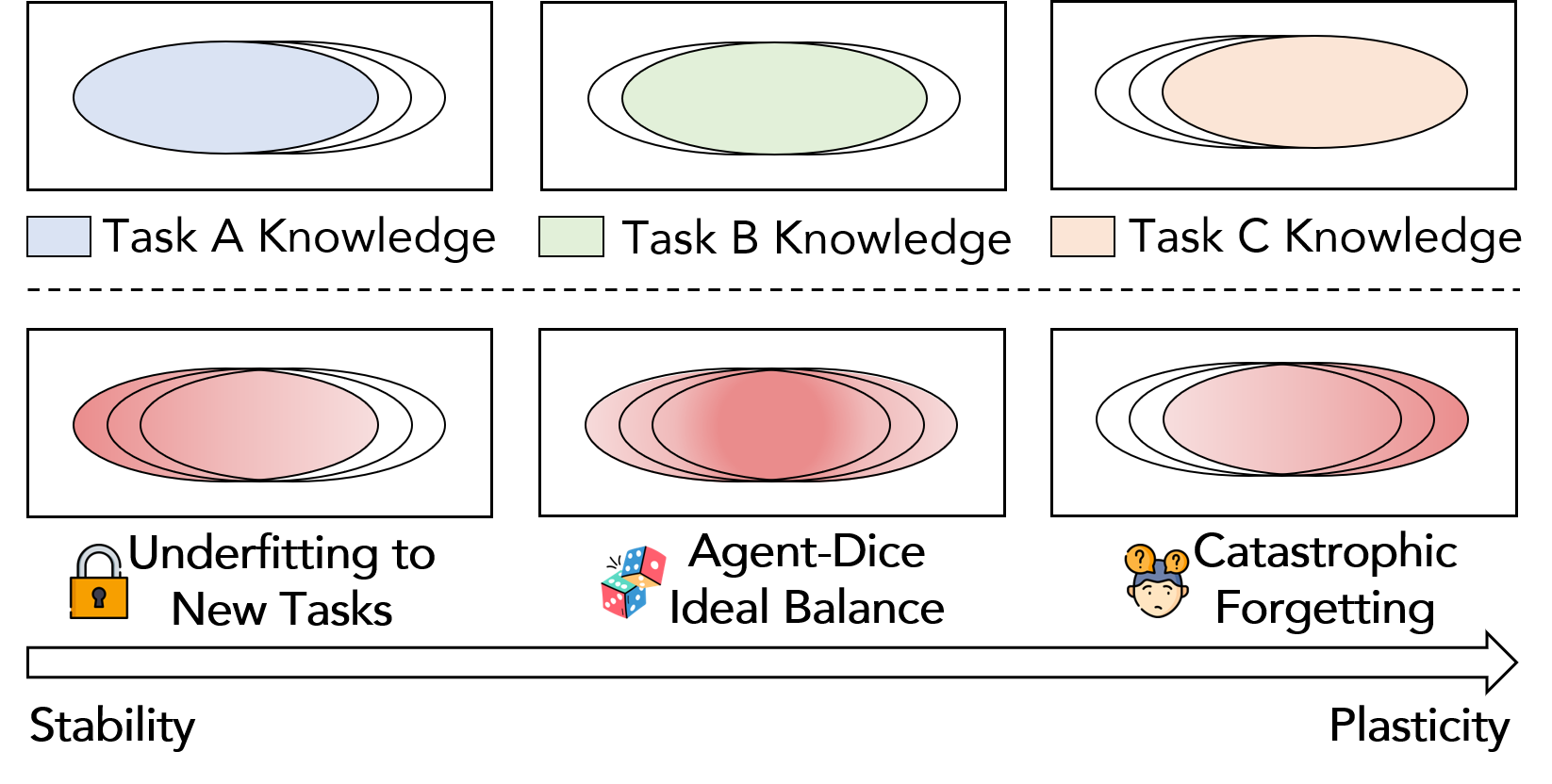}
    \caption{The stability-plasticity dilemma in agent continual learning (from Task A to Task C). Agent-Dice finds a balance between stability and plasticity by learning common knowledge.}
    \label{fig:taste}
\end{figure}

To further enhance their capabilities, it is critical for agents to possess continual learning (CL)~\cite{zheng2025lifelong,gao2025survey} skills, enabling continuous self-iteration and adaptation to novel tasks without retraining from scratch.
However, agent continual learning is fundamentally hindered by the stability-plasticity dilemma~\cite{robins1995catastrophic}. 
As shown in Figure~\ref{fig:taste}, during the adaptation process for new tasks, an agent faces a critical trade-off: overemphasizing stability hampers new learning, whereas excessive plasticity induces catastrophic forgetting.

Achieving the stability-plasticity dilemma essentially requires a precise disentanglement of knowledge updates. 
Ideally, agents are expected to minimize the interference from conflicting knowledge, while efficiently identifying and reinforcing the learning of common knowledge shared between new and old tasks to maintain plasticity.
Existing approaches primarily tackle this via two paradigms: incorporating external memory modules~\cite{ouyang2025reasoningbank,zhang2025agent} or conducting continuous iterative training~\cite{zhang2025statistical,muppidi2024fast}. However, these methods often fail to effectively distinguish between common and conflicting knowledge, inevitably leading to information loss or interference during the learning process.

To overcome these limitations, we propose Agent-Dice, a principled parameter fusion framework based on \textbf{Di}rectional \textbf{C}onsensus \textbf{E}valuation. 
Agent-Dice integrates task vectors from diverse tasks onto the original agent via a two-stage process: (i) Geometric Consensus Filtering, which prunes conflicting updates to preserve stability; and (ii) Curvature-based Importance Weighting, which amplifies shared consensus directions to enhance plasticity.


Extensive experiments in the domains of GUI agent and tool-use agent demonstrate that Agent-Dice outperforms traditional continual learning paradigms with extremely low time overhead and minimal parameter update costs.
And we further validate the rationality and effectiveness of Agent-Dice through ablation studies, model similarity analysis, and overhead evaluation.

To summarize, our contributions are four-fold:

(i) We identify that the stability–plasticity dilemma in continual learning for LLM-based agents largely arises from the failure to explicitly distinguish between common and conflicting knowledge during the learning process.

(ii) We propose Agent-Dice, a novel parameter fusion framework that integrates geometric consensus filtering with curvature-based importance weighting, enabling effective multi-task continual learning for LLM agents.

(iii) We present a theoretical analysis that proves the validity of the Agent-Dice parameter fusion scheme, while also providing new insights into the root causes of the stability–plasticity dilemma in agent continual learning.

(iv) Extensive experiments across both GUI agent and tool-use domains demonstrate that Agent-Dice exhibits outstanding continual learning performance with minimal computational overhead and parameter updates.

\section{Related Work}


In this section, we first review recent advances in LLM-based agents, focusing on two representative agent paradigms studied in this work: GUI agents and tool-use agents. 
We then summarize prior efforts on continual learning for LLMs.
Subsequently, we will then discuss progress in the field of continual learning for Agents.

\subsection{LLM Agent}
Recent advances in LLMs have empowered LLM-based agents to interact with complex environments by leveraging their capabilities reasoning~\cite{plaat2024reasoning}, planning~\cite{wei2025plangenllms}, and decision-making~\cite{sun2025llm}. 
One representative line of research focuses on GUI agents~\cite{tang2025survey,zhang2024large}, 
which operate smart devices through human-like interactions and adapt to new tasks via large-scale pre-training~\cite{wang2025ui,ye2025mobile}, 
supervised fine-tuning~\cite{ma2024coco,zhang2024you}, and reinforcement learning~\cite{tang2025gui,lu2025ui,luo2025gui,liu2025infigui,xu2025mobilerl,bai2024digirl,wangdistrl}. 
Another important direction is tool-use agents, which extend LLM capabilities by integrating external tools and APIs to perform complex reasoning and execution~\cite{ToolFormer,ToolLLM,APIGen,liutoolace,zhang2025looptool,BFCL,tau2bench,ACEBench}. 
Despite their strong performance, most agents are adapted to new domains through sequential fine-tuning or updates, which often leads to interference between previously acquired and newly learned skills. 
This limitation highlights the need for more principled continual learning mechanisms tailored to LLM agents.

\subsection{LLM Continual Learning}
To enable LLMs to better adapt to new tasks, existing studies on continual learning for LLMs have explored several main directions.
These include regularization-based methods that constrain parameter updates or feature representations~\cite{kirkpatrick2017overcoming,zenke2017continual}, approaches that store and replay a subset of previous data~\cite{rebuffi2017icarl,hou2019learning}, and architecture-based strategies that introduce task-specific modules or models~\cite{schwarz2018progress,yan2021dynamically}.
More recently, rehearsal-free methods have gained attention by leveraging parameter-efficient strategies for continual fine-tuning of pre-trained models~\cite{wang2022learning,tang2023prompt,wang2023hierarchical}.
However, most existing approaches are developed in the context of traditional LLM tasks, while continual learning for LLM agents in complex and dynamic environments presents additional challenges that have not yet been fully addressed.

\subsection{Agent Continual Learning}
The goal of agent continual learning~\cite{zheng2025lifelong,fang2025comprehensive} is to enable agents to continuously acquire new knowledge and adapt to new tasks while retaining previously learned knowledge and avoiding catastrophic forgetting.
Previous work has primarily approached agent continual learning from two perspectives: agentic reinforcement learning~\cite{zhang2025landscape} and agentic memory~\cite{xu2025mem}.
Agentic reinforcement learning adapts better to new tasks by continuously updating parameters in an online environment~\cite{zhang2025statistical,muppidi2024fast}.
Agentic memory~\cite{ouyang2025reasoningbank,zhang2025agent} adapts to new domains by continuously incorporating new knowledge into the memory module.
However, these works do not specifically adopt a paradigm design aimed at identifying common knowledge and conflicting knowledge during continual learning to mitigate the stability-plasticity dilemma.



\section{Agent-Dice}

In this section, we present \textbf{Agent-Dice}, a theoretically grounded parameter fusion framework. 
We first provide a theoretical support (complete proof provided in Appendix \ref{app:proof}) of our method using a optimization perspective. 
We then introduce the detailed implementation pipeline of Agent-Dice.
\begin{figure*}
    \centering
    \includegraphics[width=\linewidth]{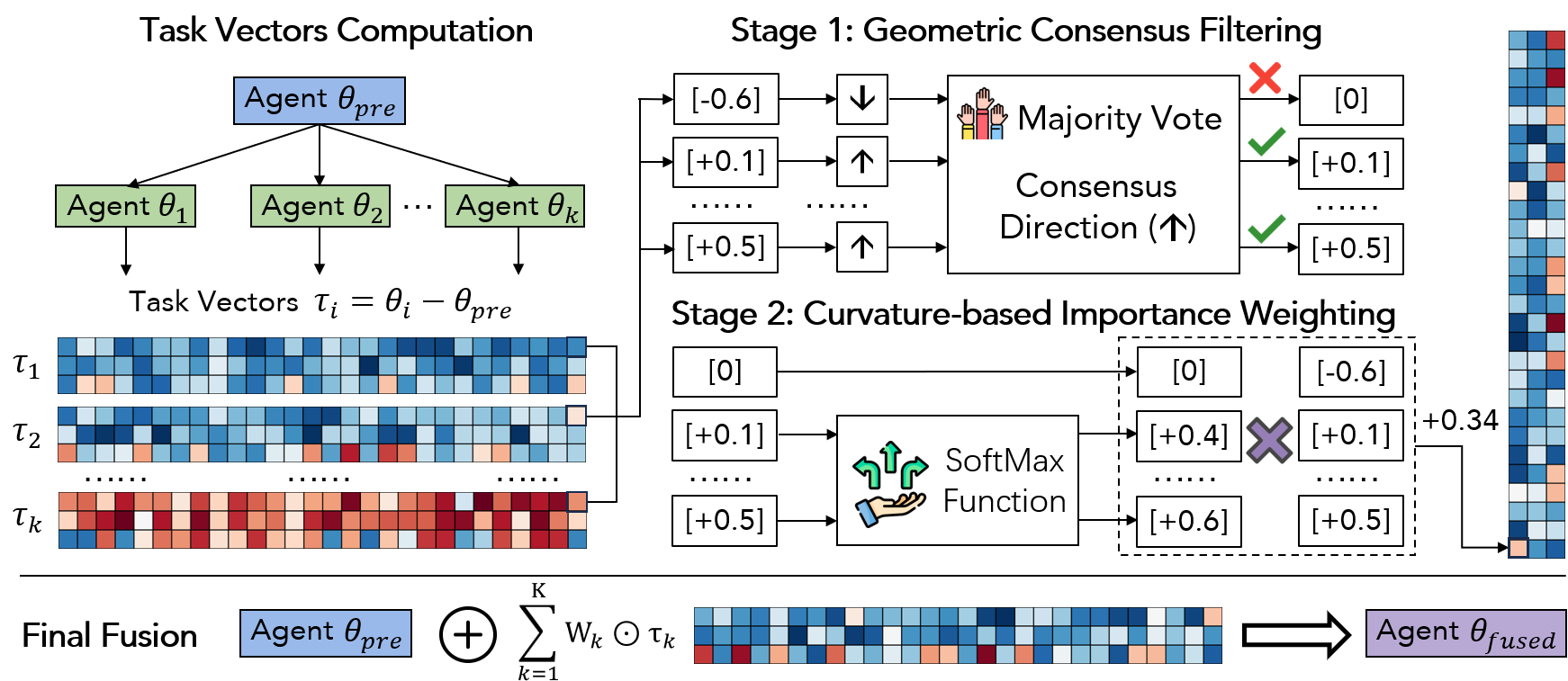}
    \caption{\textbf{The Agent-Dice Parameter Fusion Pipeline.} 
Task vectors $\boldsymbol{\tau}_k$ undergo a two-stage aggregation policy: 
\textit{Geometric Consensus Filtering} for variance reduction via outlier pruning, and 
\textit{Curvature-based Importance Weighting} for entropy maximization based on parameter saliency. 
The final refined update is added to $\boldsymbol{\theta}_{\text{pre}}$.}
    \label{fig:method}
\end{figure*}

\subsection{Motivation}
Consider an agent model with parameters $\theta_{pre} \in \mathbb{R}^d$ adapted to $K$ diverse domains $\{\mathcal{D}_k\}_{k=1}^K$ (e.g., GUI navigation and tool-use), where each adaptation yields a task-specific vector $\tau_k = \theta_k - \theta_{pre}$. We observe that the fundamental challenge in agent continual learning is the prevalence of \textit{knowledge conflict} across different domains. Formally, for any parameter dimension $i \in \{1, \dots, d\}$, there often exist domains $k$ and $j$ such that:
\begin{equation}
    \text{sgn}(\tau_{k,i}) \neq \text{sgn}(\tau_{j,i})
\end{equation}
where $\text{sgn}(\cdot)$ denotes the sign function. Such conflict indicates that the optimization directions for different agent domains are contradictory in the local parameter manifold. Standard aggregation of these updates introduces destructive interference, which manifests as the stability-plasticity dilemma and leads to catastrophic forgetting.
To bridge this gap, the agent must be able to disentangle these updates to identify a directional consensus that represents common knowledge while pruning domain-specific conflicting noise. 
This motivates the design of Agent-Dice, a fusion policy based on geometric consensus evaluation.

\subsection{Theoretical Support}

Let $\boldsymbol{\theta}^* \in \mathbb{R}^d$ be the optimal parameters on the pre-trained manifold. We consider a multi-task setting with $K$ tasks, where each task $k$ is associated with a loss function $\mathcal{L}_k: \mathbb{R}^d \to \mathbb{R}$. The fine-tuned parameter vector for task $k$ is denoted by $\boldsymbol{\theta}_k = \boldsymbol{\theta}_{\text{pre}} + \boldsymbol{\tau}_k$, where $\boldsymbol{\tau}_k$ represents the task-specific displacement vector.

Our goal is to find a fusion policy $\Phi$ such that the fused parameter $\boldsymbol{\theta}_{\text{fused}} = \Phi(\{\boldsymbol{\theta}_k\}_{k=1}^K)$ minimizes the worst-case approximation error relative to the Pareto-optimal solution of the joint loss $\mathcal{L}_{\text{total}}(\boldsymbol{\theta}) = \sum_{k=1}^K \mathcal{L}_k(\boldsymbol{\theta})$. We analyze the fusion process through three theoretical lenses: linear approximation, variance reduction via consensus, and maximum entropy weight assignment.

\paragraph{Parameter Space Linearization.}
First, we establish the validity of the linear combination form applied in an element-wise manner (meaning each parameter is combined individually).
We rely on the assumption that the pre-trained model lies in a \textit{linear mode connectivity} basin~\cite{mirzadehlinear}, a phenomenon widely observed in large-scale deep learning models.
\begin{thmbox}
\begin{theorem}[First-Order Manifold Aggregation]
Assume that for a local neighborhood around $\boldsymbol{\theta}_{\text{pre}}$, the loss function $\mathcal{L}_k$ is approximately linear with respect to $\boldsymbol{\tau}_k$. 
Let $\mathbf{W} \in \mathbb{R}^{d \times K}$ be a weighting matrix where $\sum_{k=1}^K w_{k,i} = 1$. 
The update rule $\boldsymbol{\theta}_{\text{new}} = \boldsymbol{\theta}_{\text{pre}} + \sum_{k=1}^K \mathbf{w}_k \odot \boldsymbol{\tau}_k$ approximates a single gradient descent step on a surrogate multi-task objective $\tilde{\mathcal{L}}(\boldsymbol{\theta}) = \sum_{k=1}^K \mathbf{w}_k^\top \mathcal{L}_k(\boldsymbol{\theta})$.
\end{theorem}
\end{thmbox}

\begin{proof}
Using a first-order Taylor expansion, $\mathcal{L}_k(\boldsymbol{\theta}_{\text{pre}} + \boldsymbol{\tau}) \approx \mathcal{L}_k(\boldsymbol{\theta}_{\text{pre}}) + \nabla \mathcal{L}_k(\boldsymbol{\theta}_{\text{pre}})^\top \boldsymbol{\tau}$. Since $\boldsymbol{\tau}_k$ is obtained via SGD, $\boldsymbol{\tau}_k \propto -\nabla \mathcal{L}_k(\boldsymbol{\theta}_{\text{pre}})$. 
The fused update becomes:
$\Delta \boldsymbol{\theta} \propto -\nabla \left( \sum_{k=1}^K \mathbf{w}_k^\top \mathcal{L}_k(\boldsymbol{\theta}_{\text{pre}}) \right).$
This confirms that the fusion rule minimizes the joint loss.
\end{proof}

\paragraph{Noise Suppression via Geometric Consistency.}
Multi-task fusion often suffers from gradient interference. 
We model the task vectors as noisy estimators of a shared latent descent direction to justify the necessity of consensus-based filtering.

\begin{defbox}
\textbf{Definition (Interference Model).} For a parameter $j$, let the true descent sign be $s^*_j \in \{-1, +1\}$. We assume that the sign of the $k$-th task update, $s_{k,j} = \text{sgn}(\tau_{k,j})$, follows a Bernoulli distribution with success probability $p > 0.5$, i.e., $P(s_{k,j} = s^*_j) = p$.
\end{defbox}

\begin{thmbox}
\begin{theorem}[Consensus-Induced Variance Reduction]
Let $\mathcal{S}_j$ be the set of tasks with consistent signs for parameter $j$. If outlier tasks (where $s_{k,j} \neq s^*_j$) are excluded from aggregation, the probability of update error decays exponentially with the size of the consensus set $|\mathcal{S}_j|$, strictly outperforming standard averaging.
\end{theorem}
\end{thmbox}

\begin{proof}
Let $X$ be the number of consistent tasks. By Hoeffding's inequality, the probability that the majority vote is incorrect is bounded by:
\begin{equation}
P(\text{error}) \le \exp(-2|\mathcal{S}_j|(p-0.5)^2).    
\end{equation}

Standard averaging includes the minority set, which effectively reduces the margin $p-0.5$ or introduces destructive interference, thus increasing the error bound. Filtering ensures that the update remains within the cone of the true gradient.
\end{proof}

\paragraph{Saliency Maximization via Boltzmann Distribution.}

Finally, we formalize the assignment of scalar weights.
We posit that the magnitude $|\tau_{k,j}|$ serves as a proxy for the local sensitivity (curvature) of the loss landscape, and thus cast the weight selection as a Maximum Entropy problem.

\begin{thmbox}
\begin{theorem}[Optimal Weighting under Saliency Constraints]
Let $u_{k,j} = |\tau_{k,j}|$ denote the saliency of parameter $j$ for task $k$. The probability distribution $\mathbf{w}_j$ that maximizes the entropy $H(\mathbf{w}_j)$ subject to the constraint of matching the expected saliency is the Boltzmann distribution:
\begin{equation}
w_{k,j} = \frac{\exp(\beta u_{k,j})}{\sum_{m \in \mathcal{S}_j} \exp(\beta u_{m,j})},
\end{equation}

where $\beta$ is the inverse temperature parameter.
\end{theorem}
\end{thmbox}

\begin{proof}
We formulate the Lagrangian $\mathcal{L} = -\sum_k w_{k,j} \log w_{k,j} + \lambda (\sum_k w_{k,j} u_{k,j} - C) + \gamma (\sum_k w_{k,j} - 1)$. 
Setting the partial derivative $\frac{\partial \mathcal{L}}{\partial w_{k,j}} = 0$ yields $\log w_{k,j} = \lambda u_{k,j} + \gamma - 1$, which implies $w_{k,j} \propto \exp(\lambda u_{k,j})$. This derivation confirms that Softmax is the least biased distribution given the saliency magnitudes.
\end{proof}

\subsection{Method: Directional Consensus Evaluation for Parameter Fusion}

Guided by the theoretical support, we formalize the Agent-Dice algorithm. 
The method operates element-wise on the agent parameters to construct a fused update. 
To provide a clear overview, we first present the \textit{general formulation} of the fusion process. 
Supported by Theorem 1, the final fused parameter vector $\boldsymbol{\theta}_{\text{fused}}$ is obtained by aggregating the candidate task vectors $\{\boldsymbol{\tau}_1, \dots, \boldsymbol{\tau}_K\}$ weighted by a dynamic consensus matrix $\mathbf{W}_k$:
\begin{equation}
    \boldsymbol{\theta}_{\text{fused}} = \boldsymbol{\theta}_{\text{pre}} + \sum_{k=1}^K \mathbf{W}_k \odot \boldsymbol{\tau}_k,
\label{equation:main}
\end{equation}
where $\odot$ denotes the Hadamard product (element-wise multiplication), and $\mathbf{W}_k \in \mathbb{R}^d$ represents the element-wise importance weight vector associated with the agent parameters $\boldsymbol{\theta}_k$ learned from the $k$-th task.
The core innovation of Agent-Dice lies in the specific construction of $\mathbf{W}_k$, which is determined through a two-stage process: geometric consensus filtering and curvature-based importance weighting for continual learning.

\paragraph{Stage 1: Geometric Consensus Filtering.}
Supported by Theorem 2, we first identify the dominant optimization direction to construct the active set for each parameter. 
This step acts as a binary mask, filtering out outlier updates that contradict the manifold consensus.

Let $\tau_{k,i}$ denote the update value of the $k$-th agent for the $i$-th parameter, where $i \in \{1, \dots, d\}$. We define the sign indicator $s_{k,i} \in \{0,1\}$ as $s_{k,i} = 1(\tau_{k,i} \ge 0)$.
The consensus score for the $i$-th parameter is given by the positive vote count $V_i = \sum_{k=1}^K s_{k,i}$. 
The active set $\mathcal{S}_i$ for this specific parameter is determined by a majority threshold $\delta$ (in this paper, $\delta$ is set to $K/2$):
\begin{equation}
    \mathcal{S}_i = 
    \begin{cases} 
        \{k \mid s_{k,i} = 1\}, & \text{if } V_i > \delta,  \\
        \{k \mid s_{k,i} = 0\}, & \text{if } V_i < K - \delta, \\
        \{1, \dots, K\}, & \text{otherwise}. \quad 
    \end{cases}
\end{equation}

Only agents belonging to $\mathcal{S}_i$ are considered eligible to contribute to the fusion of the $i$-th parameter; effectively, weights for agents $k \notin \mathcal{S}_i$ will be forced to zero.

\paragraph{Stage 2: Curvature-based Importance Weighting.}
After filtering, supported by Theorem 3, we determine the specific values of the weights $w_{k,i}\in\mathbf{W}_k$ for the eligible agents. 
Large update magnitudes $|\tau_{k,i}|$ typically indicate high confidence or traversal through steep gradients in the loss landscape (high curvature) for that specific parameter. To prioritize these informative features, we employ a masked Softmax function to normalize these magnitudes within the active set for each parameter $i$:
\begin{equation}
    w_{k,i} = 
    \begin{cases} 
        \frac{\exp(|\tau_{k,i}|)}{\sum_{j \in \mathcal{S}_i} \exp(|\tau_{j,i}|)}, & \text{if } k \in \mathcal{S}_i, \\
        0, & \text{if } k \notin \mathcal{S}_i.
    \end{cases}
\end{equation}

This weighting scheme ensures that the final model trajectory follows the consensus of the most confident agents locally for each parameter, effectively neutralizing catastrophic forgetting caused by conflicting tasks.

\paragraph{Final Fusion.}
By substituting the computed element-wise weights $w_{k,i}$ back into the general formulation in Equation~\ref{equation:main}, we obtain the final updated parameter vector $\boldsymbol{\theta}_{\text{fused}}$. 
This aggregation effectively integrates the directional consensus with magnitude-based confidence. 
Specifically, for every parameter $i$, the update becomes a weighted sum $\sum_{k \in \mathcal{S}_i} w_{k,i} \tau_{k,i}$, where the contribution of conflicting agents is nullified. 
Consequently, the fused model updates strictly along the manifold direction determined by the majority, while the step size is adaptively governed by the agents exhibiting the strongest local feature response. 
This ensures the global optimization trajectory balances stability and plasticity via consensus filtering and curvature weighting, respectively.

\begin{table*}[t]
\centering
\footnotesize 
\setlength{\tabcolsep}{3.5pt} 
\renewcommand{\arraystretch}{1.1}

\begin{tabular}{l c c c c c c c c c c}
\toprule
\multirow{3}{*}{Method} & 
\multicolumn{3}{c}{\textbf{AITZ}} & 
\multicolumn{3}{c}{\textbf{AndroidControl}} & 
\multicolumn{3}{c}{\textbf{GUI-Odyssey}} & 
\multirow{2}{*}{AvgZ}
 \\
\cmidrule(lr){2-4} \cmidrule(lr){5-7} \cmidrule(lr){8-10}
& Type & SR & TSR & 
Type & SR & TSR & 
Type & SR & TSR \\
\midrule

\textbf{Zero-Shot} & 
63.41 & \uwave{45.08} & \uwave{0.20} & 
73.12 & 47.14 & 13.63 &
74.22 &54.71 &0.60 & -0.38\\
\midrule

\textbf{Learn from AITZ} & 
\textbf{75.63} & \textbf{59.92} & \textbf{6.13}  & 
61.82 & 36.70 & 7.54 & 
83.33 & 60.19 & 0.54 & \uwave{0.09} \\

\textbf{Learn from AndroidControl} & 
61.34 & 30.85 & 0.00 & 
\textbf{85.25} & \textbf{57.95} & \textbf{20.31} & 
74.68 & 40.80 & 0.24 & -0.26 \\

\textbf{Learn from GUI-Odyssey} & 
63.16 & 37.67 & \uwave{0.20} & 
69.03 & 36.79 & 6.29 & 
\textbf{90.74} & \textbf{76.06} & \textbf{4.62} & -0.17\\

\midrule

\textbf{CL from AITZ and AndroidControl} & 
\uwave{65.81} & 37.41 & 0.00 & 
\underline{84.49} & \underline{57.47} & \underline{19.40} & 
73.85 & 39.95 & 0.30 & -0.14\\

\textbf{CL from all three} & 
65.78 & 42.05 & 0.04 & 
73.56 & 43.48 & 9.57 & 
\underline{90.69} & \underline{75.79} & \underline{4.44} & \underline{0.14} \\

\midrule
\rowcolor{blue!5}
\textbf{Agent-Dice (Ours)} & 
\underline{74.72} & \underline{57.10} & \underline{2.37} & 
\uwave{80.03} & \uwave{51.42} & \uwave{14.42} & 
\uwave{89.27} & \uwave{72.28} & \uwave{2.10} & \textbf{0.73}
\\
\bottomrule
\end{tabular}
\caption{Experiment results of Agent-Dice in the GUI agent domain, with OS-Atlas-Pro-7B as the base model. The best results are highlighted in \textbf{bold}, while the second-best are \underline{underlined} and the third-best are \uwave{underwaved}.}
\label{tab:GUI_atlas_results}
\end{table*}

\begin{table*}[t]
\centering
\footnotesize
\setlength{\tabcolsep}{3.5pt}
\renewcommand{\arraystretch}{1.1}

\begin{tabular}{l c c c c c c c c c c}
\toprule
\multirow{3}{*}{Method} & 
\multicolumn{3}{c}{\textbf{AITZ}} & 
\multicolumn{3}{c}{\textbf{AndroidControl}} & 
\multicolumn{3}{c}{\textbf{GUI-Odyssey}} &
\multirow{2}{*}{AvgZ}
 \\
\cmidrule(lr){2-4} \cmidrule(lr){5-7} \cmidrule(lr){8-10}
& Type & SR & TSR & 
Type & SR & TSR & 
Type & SR & TSR \\
\midrule

\textbf{Zero-Shot} & 
56.81 & \uwave{41.14} & \underline{0.99} & 
73.90 & \uwave{52.18} & \uwave{13.76} &
67.68 & 44.08 & \uwave{0.42} & -0.02 \\
\midrule

\textbf{Learn from AITZ} & 
\textbf{74.72} & \textbf{58.23} & \textbf{5.34} & 
58.00 & 30.23 & 3.80 &
66.56 & 37.17 & 0.00 & -0.03\\

\textbf{Learn from AndroidControl} & 
49.12 & 20.56 & 0.00 & 
\textbf{83.21} & \textbf{61.92} & \textbf{20.12} &
65.22 & 28.04 & 0.00 & -0.30\\

\textbf{Learn from GUI-Odyssey} & 
58.06 & 32.01 & 0.20 & 
66.78 & 31.07 & 4.72 &
\textbf{88.54} & \textbf{67.42} & \textbf{1.86} &\underline{}{0.10}\\

\midrule

\textbf{CL from AITZ and AndroidControl} & 
59.45 & 31.44 & 0.00 & 
\underline{82.90} & \underline{61.91} & \underline{18.94} &
64.99 & 28.66 & 0.00 & -0.06\\

\textbf{CL from all three} & 
\uwave{60.03} & 35.97 & \uwave{0.59} & 
72.47 & 34.23 & 5.05 &
\underline{88.09} & \underline{63.83} & \underline{0.48} &\uwave{0.01}\\

\midrule
\rowcolor{blue!5}
\textbf{Agent-Dice (Ours)} & 
\underline{68.28} & \underline{47.49} & 0.40 & 
\uwave{79.39} & 41.50 & 8.45 &
\uwave{81.40} & \uwave{54.27} & \uwave{0.42} & \textbf{0.29}\\
\bottomrule
\end{tabular}
\caption{Experiment results of Agent-Dice in the GUI agent domain, with Qwen3-VL-8B as the base model. The best results are highlighted in \textbf{bold}, while the second-best are \underline{underlined} and the third-best are \uwave{underwaved}.}
\label{tab:GUI_qwen3vl_results}
\end{table*}

\begin{table*}[t]
\centering
\footnotesize
\setlength{\tabcolsep}{4.5pt}
\renewcommand{\arraystretch}{1.1}

\begin{tabular}{l c c c c c c c c c}
\toprule
\multirow{3}{*}{Method} &
\multicolumn{2}{c}{\textbf{Subset 0}} &
\multicolumn{2}{c}{\textbf{Subset 1}} &
\multicolumn{2}{c}{\textbf{Subset 2}} &
\multicolumn{2}{c}{\textbf{Subset 3}} &
\multirow{2}{*}{AvgZ}
\\
\cmidrule(lr){2-3} \cmidrule(lr){4-5} \cmidrule(lr){6-7} \cmidrule(lr){8-9}
& Func & Full &
Func & Full &
Func & Full &
Func & Full \\
\midrule

\textbf{Zero-Shot} &
\textbf{99.64} & 81.85 &
\underline{99.26} & 85.66 &
98.52 & 83.85 &
\uwave{99.28} & 86.36 & -1.42 \\
\midrule
\textbf{Learn from Subset 0} &
\uwave{98.93} & 85.96 &
\textbf{99.63} & 88.83 &
\underline{99.26} & \uwave{88.81} &
\textbf{100.0} & \uwave{91.35} & 0.27\\

\textbf{Learn from Subset 1} &
\underline{99.29} & \uwave{86.82} &
\textbf{99.63} & 87.52 &
\uwave{98.89} & \underline{89.36} &
\underline{99.64} & 90.52 & 0.13\\

\textbf{Learn from Subset 2} &
98.57 & 84.59 &
\textbf{99.63} & 87.71 &
97.79 & 87.52 &
98.57 & 90.35 &-1.02\\

\textbf{Learn from Subset 3} &
\textbf{99.64} & 85.79 &
\textbf{99.63} & 88.83 &
\underline{99.26} & 88.62 &
\underline{99.64} & 89.85 & 0.28\\
\midrule
\textbf{CL from Subset 0 \& 1} &
\underline{99.29} & 86.30 &
\textbf{99.63} & \underline{89.57} &
\underline{99.26} & \textbf{89.54} &
\underline{99.64} & 91.18 & \uwave{0.44}\\

\textbf{CL from Subset 0, 1 \& 2} &
\underline{99.29} & \underline{87.16} &
\textbf{99.63} & \uwave{89.01} &
\underline{99.26} & 88.07 &
\textbf{100.0} & \underline{91.68} & \underline{0.48}\\

\textbf{CL from all Subsets} &
\textbf{99.64} & 86.13 &
\uwave{98.88} & 88.64 &
\textbf{99.63} & 88.44 &
\underline{99.64} & 90.68 &0.06\\

\midrule
\rowcolor{blue!5}
\textbf{Agent-Dice (Ours)} &
\underline{99.29} & \textbf{87.33} &
\textbf{99.63} & \textbf{90.69} &
\underline{99.26} & \underline{89.36} &
\textbf{100.0} & \textbf{92.18} & \textbf{0.79}\\
\bottomrule
\end{tabular}
\caption{Experiment results of Agent-Dice in the tool-use domain, with Qwen3-8B as the base model. The best results are highlighted in \textbf{bold}, while the second-best are \underline{underlined} and the third-best are \uwave{underwaved}.}
\label{tab:tool_qwen3_results}
\end{table*}

\begin{table*}[t]
\centering
\footnotesize
\setlength{\tabcolsep}{4.5pt}
\renewcommand{\arraystretch}{1.1}

\begin{tabular}{l c c c c c c c c c}
\toprule
\multirow{3}{*}{Method} &
\multicolumn{2}{c}{\textbf{Subset 0}} &
\multicolumn{2}{c}{\textbf{Subset 1}} &
\multicolumn{2}{c}{\textbf{Subset 2}} &
\multicolumn{2}{c}{\textbf{Subset 3}} &
\multirow{2}{*}{AvgZ}\\
\cmidrule(lr){2-3} \cmidrule(lr){4-5} \cmidrule(lr){6-7} \cmidrule(lr){8-9}
& Func & Full &
Func & Full &
Func & Full &
Func & Full \\
\midrule

\textbf{Zero-Shot} &
13.93 & 5.82 &
14.87 & 6.89 &
13.28 & 5.14 &
10.49 & 10.32 &-2.81\\
\midrule
\textbf{Learn from Subset 0} &
\textbf{98.57} & \uwave{79.45} &
\uwave{97.77} & 82.31 &
\textbf{98.52} & \underline{83.67} &
\textbf{98.57} & \underline{85.19} & \underline{0.45}\\

\textbf{Learn from Subset 1} &
88.57 & 72.95 &
91.45 & 73.37 &
92.62 & 75.05 &
89.96 & 75.87 & 0.13\\

\textbf{Learn from Subset 2} &
\uwave{96.79} & 74.14 &
96.65 & 76.54 &
\underline{97.05} & 74.50 &
96.77 & 80.37 & 0.29\\

\textbf{Learn from Subset 3} &
95.36 & 78.25 &
\textbf{98.88} & \uwave{83.05} &
\underline{97.05} & 80.55 &
95.34 & \uwave{83.86} & 0.39\\
\midrule
\textbf{CL from Subset 0 \& 1} &
95.71 & \underline{79.62} &
97.40 & \underline{83.24} &
\uwave{96.31} & \uwave{81.10} &
96.42 & 83.19 & \uwave{0.40} \\

\textbf{CL from Subset 0, 1 \& 2} &
93.57 & 74.14 &
93.31 & 76.72 &
93.73 & 78.53 &
93.91 & 82.20 & 0.26\\

\textbf{CL from all Subsets} &
95.00 & 78.94 &
94.42 & 82.50 &
95.94 & \uwave{81.10} &
\uwave{97.49} & 83.19 & 0.38\\

\midrule
\rowcolor{blue!5}
\textbf{Agent-Dice (Ours)} &
\underline{97.50} & \textbf{84.25} &
\underline{98.14} & \textbf{85.47} &
\textbf{98.52} & \textbf{85.69} &
\underline{98.21} & \textbf{87.02} & \textbf{0.51} \\
\bottomrule
\end{tabular}
\caption{Experiment results of Agent-Dice in the tool-use domain, with Llama-3.1-8B as the base model. The best results are highlighted in \textbf{bold}, while the second-best are \underline{underlined} and the third-best are \uwave{underwaved}.}
\label{tab:tool_llama3_results}
\end{table*}

\section{Experiments}
In this section, we validate the effectiveness of Agent-Dice in two domains: GUI agent and tool-use agent.
First, we will briefly introduce the implementation, and then present our and analyse our main results.

\subsection{Implementation}
\paragraph{Dataset.} For the GUI agent domain, we choose three popular benchmarks: AITZ~\cite{zhang2024android}, AndroidControl~\cite{li2024effects}, and GUI-Odyssey~\cite{lu2025guiodyssey}. 
For the tool-use agent domain, we chose ToolACE~\cite{liutoolace} as the dataset and partitioned it into four subsets based on the greedy algorithm (Appendix~\ref{app:toolace}) according to the minimum tool overlap. 

\paragraph{Evaluation Protocol.} We simulate the learning paradigm of continual learning agents by incrementally adding new knowledge to supervised fine-tuning the agent. 
For the GUI agent domain, we add new benchmarks to train the agent. 
For the tool-use agent domain, we gradually add new subsets to train the agent.
We report the results in the zero-shot setting, the setting where each task is trained individually, and the setting where tasks are trained sequentially and continuously.
More detailed evaluation protocol can be found in Appendix~\ref{app:experiment}.

\paragraph{Metrics.} Our core reported metric is the average Z-score (AvgZ) from multi-task learning evaluation. 
Let \( \{M_i\}_{i=1}^{N} \) be the metric scores across \(N\) tasks. We compute:
\(
\text{Z-score}(M_i) = \frac{M_i - \mu_i}{\sigma_i},
\quad
\text{AvgZ} = \frac{1}{N}\sum_{i=1}^{N} \text{Z-score}(M_i)
\).
where \( \mu_i \) and \( \sigma_i \) are the mean and standard deviation from baseline models on task \(i\).

\begin{itemize}
    \item For the GUI agent domain, we also report the action type accuracy (Type), step-wise success rate (SR), and trajectory success rate (TSR), where TSR equals 1 only if the SR for every step in the trajectory is 1.
    \item For the tool-use agent domain, we report the rate of predicting the correct tool function name (Func) and the rate of correctly predicting both the tool function name and its parameters (Full).
\end{itemize}

\paragraph{Models.} For the GUI agent domain, we select OS-Atlas-Pro-7B~\cite{wuatlas} and Qwen3-VL-8B~\cite{Qwen3-VL} for experimentation. 
This is because OS-Atlas-Pro-7B is a model specifically designed for the GUI agent domain, while Qwen3-VL-8B is a general-purpose model that has undergone training in the GUI agent domain, making both representative choices. 
For the tool-use agent domain, we select Qwen3-8B~\cite{qwen3} and Llama-3.1-8B~\cite{dubey2024llama} for experimentation. 
Qwen3-8B inherently possesses tool-use capabilities, allowing it to achieve good tool-use performance in a zero-shot setting, whereas Llama-3.1-8B lacks tool-use capabilities. 
Thus, these two models represent distinct scenarios.

\paragraph{Implementation Details.}
All experiments were performed using 1200 hours of 80GB GPU computing resources. 
We conducted training with llama-factory, setting a learning rate of 1e-5 for 3 epochs when using OS-Atlas-pro-7B as the base model in the GUI agent domain, and a learning rate of 1e-5 for 2 epochs when using Qwen3-VL-8B as the base. 
For the tool-use agent domain, a learning rate of 1.0e-5 was applied for 3 epochs when using both Qwen3-8B and Llama-3.1-8B as base models.

\subsection{Main Results}

The experimental results, as shown in Tables~\ref{tab:GUI_atlas_results}-\ref{tab:tool_llama3_results}, lead to the following key findings:

(i) In both the GUI agent domain and the agent tool‑use domain, Agent‑Dice achieves the highest AvgZ. 
This indicates that Agent‑Dice outperforms traditional lifelong‑learning agents in incremental learning, demonstrating its effectiveness.

(ii) Overall, the AvgZ of the zero-shot setting in each experimental group is negative, indicating that training can indeed enhance the overall capability of the agent.
Moreover, the learn from all setting is often not the second-best, suggesting that when continuously learning new knowledge, the agent's old knowledge may be affected, leading to a decline in overall capability.

(iii) For the GUI agent domain, due to the significant differences in APP data across different datasets, a clear catastrophic forgetting problem is observed. 
When a agent encounters knowledge from GUI‑Odyssey, it exhibits noticeable forgetting of the knowledge related to AITZ and Androidcontrol APPs. 
With Agent‑Dice, compared to learning from all three datasets, the performance on AITZ and Androidcontrol is greatly improved while only a slight decline in metrics is observed on GUI‑Odyssey.


(iv) For the agent tool‑use domain, Agent‑DiCE yields more pronounced gains via common knowledge reinforcement and noise filtering, given minor cross-tool learning mechanism disparities in those tasks. It achieves top metrics across nearly all subsets and works effectively for both tool-use-capable (Qwen3-8B) and tool-use-less-capable (Llama-3.1-8B) models.

\begin{figure*}[htbp]
\centering
\begin{minipage}[t]{0.45\textwidth}
\centering
\includegraphics[width=\textwidth]{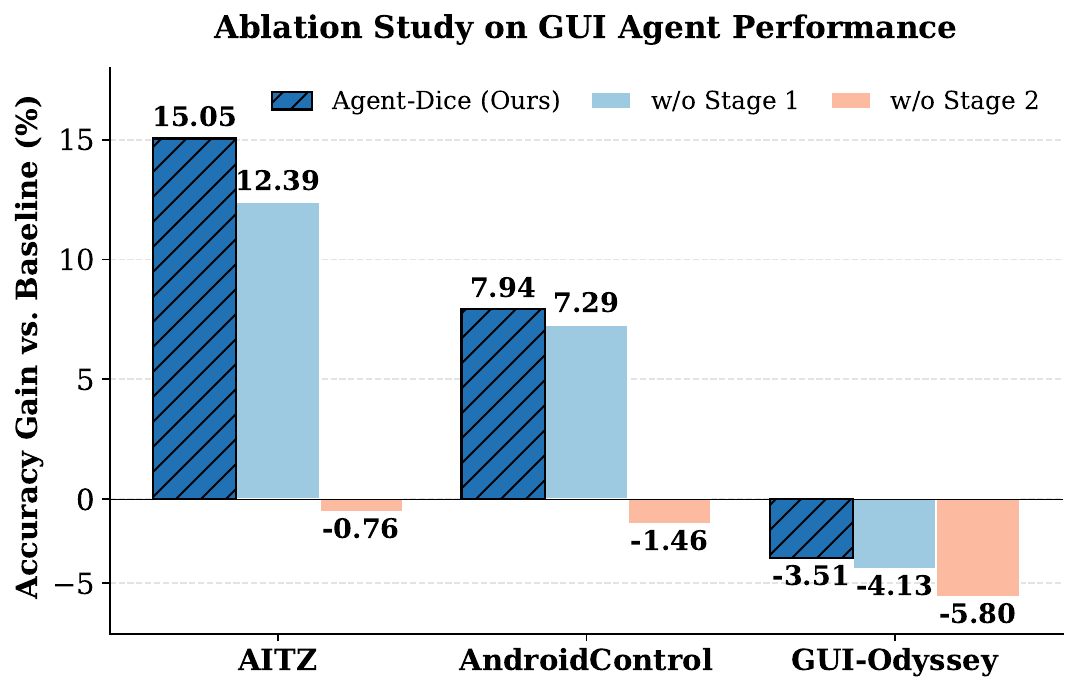}
\caption{Ablation study on GUI Agent tasks.}
\label{fig:ablation_gui}
\end{minipage}
\begin{minipage}[t]{0.45\textwidth}
\centering
\includegraphics[width=\textwidth]{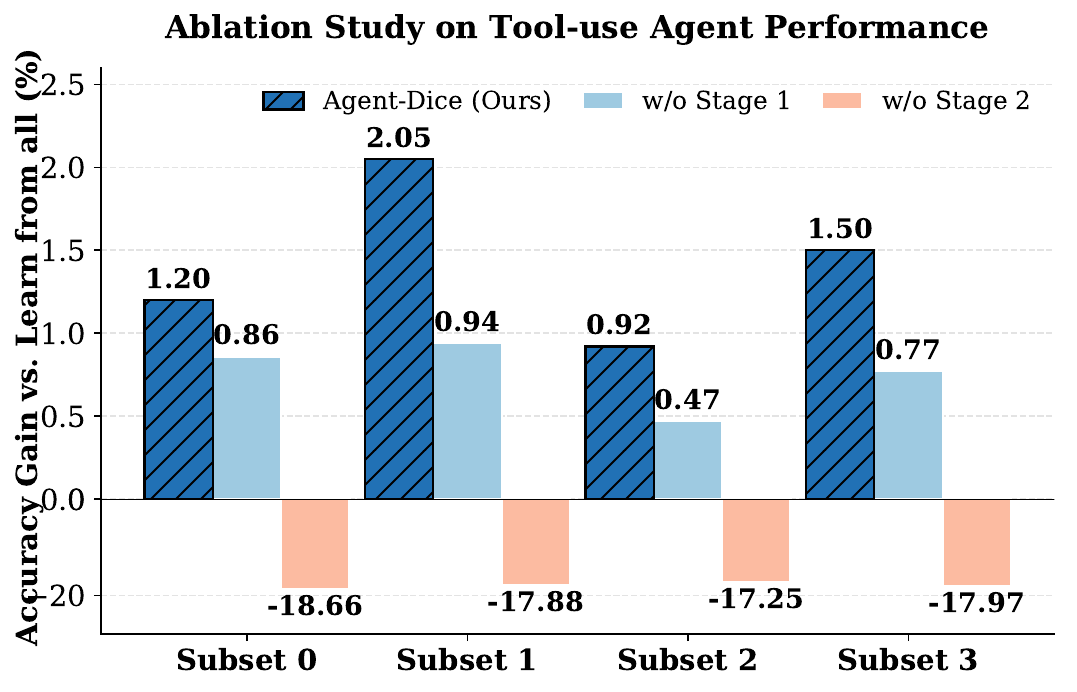}
\caption{Ablation study on Tool-use Agent tasks.}
 \label{fig:ablation_tool}
\end{minipage}
\vspace{-6mm}
\end{figure*}


\section{Further Analysis}
In this section, we first validate the rationale of Agent-Dice’s design through an ablation study. 
Then, we differentiate Agent-Dice from direct sequential learning of task knowledge via a model similarity analysis. 
Finally, we demonstrate the lightweight nature of Agent-Dice through an overhead evaluation.

\subsection{Ablation Study}

We conduct an ablation study on the two stages of Agent-Dice: geometric consensus filtering and curvature-based importance weighting. For stage 1 ablation, we remove the voting-based weighting mechanism and instead assign uniform weights to all new task vectors during training.
For stage 2 ablation, we discard the final importance weighting step, i.e., no curvature-based reweighting is applied to the updates. We report the Full metric on the tool-use agent domain and the SR metric on the GUI agent domain.

Results are illustrated in Figures ~\ref{fig:ablation_gui}-\ref{fig:ablation_tool}. Under the stage 1 ablation setting, the agent fails to learn precise common knowledge across tasks, highlighting the critical role of geometric consensus filtering.
Under the stage 2 ablation setting, the agent’s performance drops sharply in both the GUI agent and tool-use agent domains. 
This degradation is caused by the absence of curvature-based importance weighting, which otherwise constrains the update magnitude; without this constraint, the knowledge updates become excessively large and unstable. Overall, these results demonstrate that both stages are indispensable for stable and effective knowledge integration in Agent-Dice.

\begin{figure}[t]
    \centering
    \includegraphics[width=\linewidth]{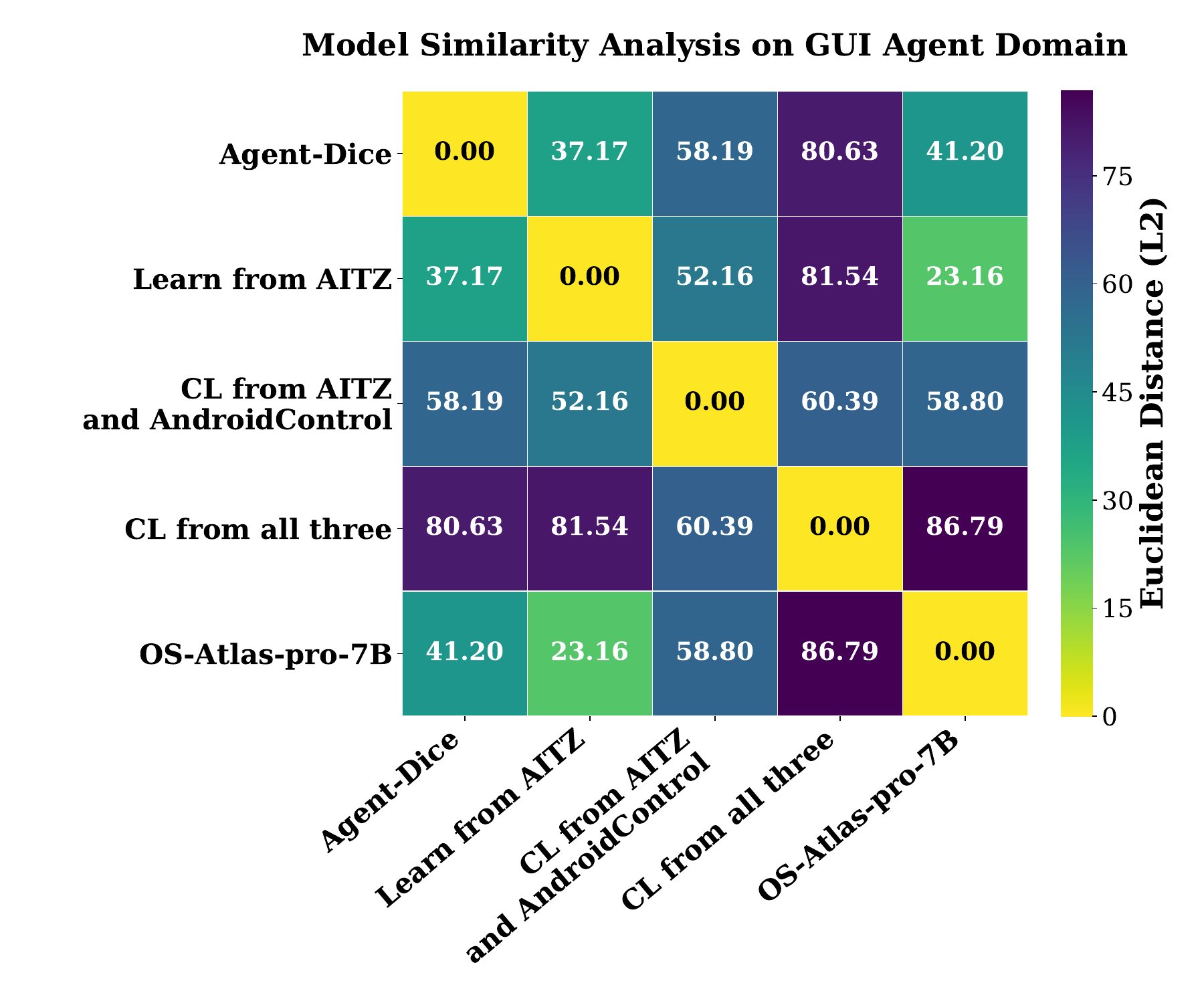}
    \vspace{-9mm}
    \caption{Model similarity analysis in the GUI agent domain. The similarity between Agent-Dice and OS-Atlas-Pro-7B is only marginally lower than that of models trained on a single dataset.}
    \label{fig:similar_atlas}
    \vspace{-2mm}
\end{figure}

\begin{figure}[t]
    \centering
    \includegraphics[width=\linewidth]{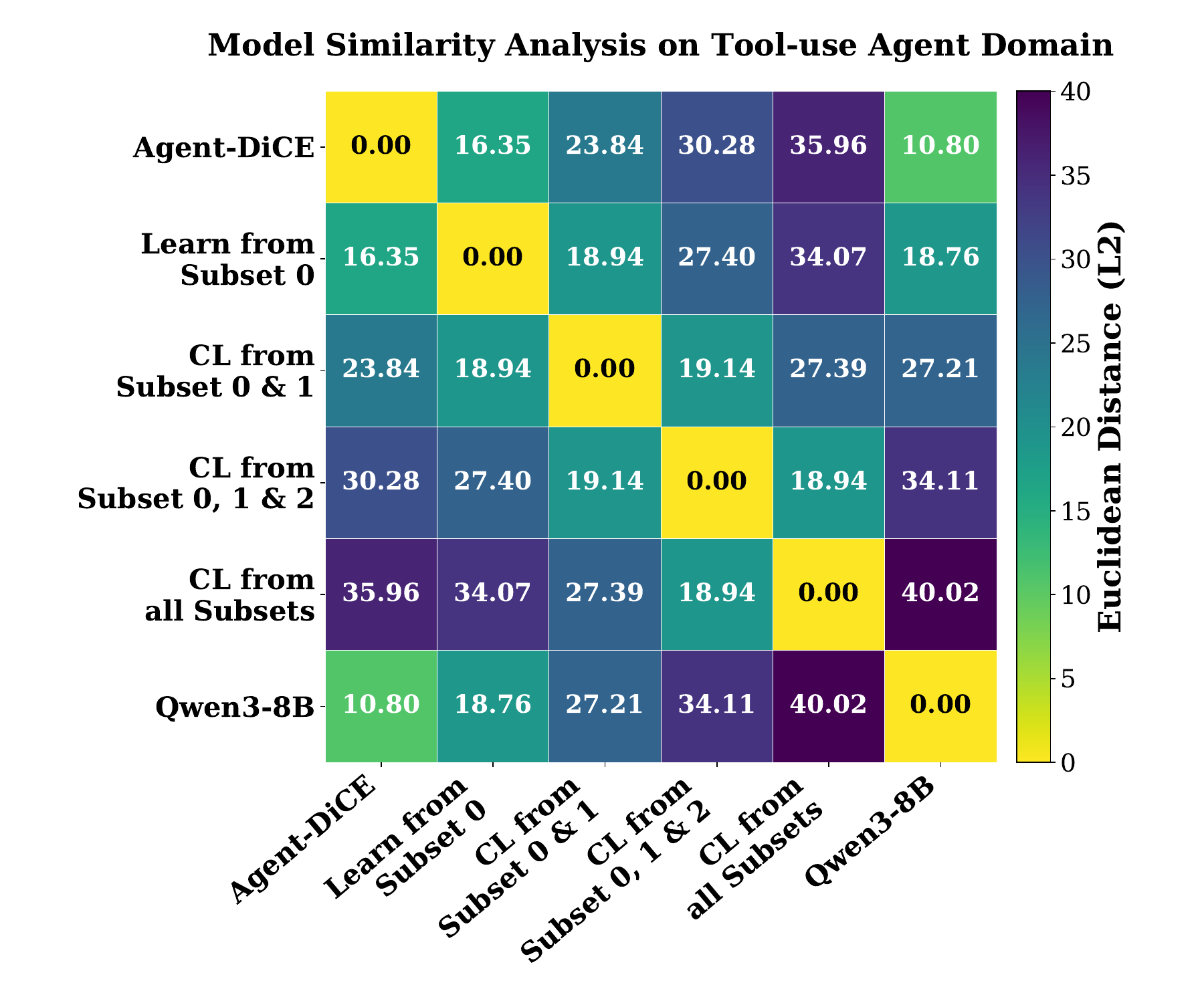}
    \vspace{-9mm}
    \caption{Model similarity analysis in the tool-use agent domain. Agent-Dice exhibits the highest similarity to Qwen3-8B.}
    \label{fig:similar_qwen3}
    \vspace{-2mm}
\end{figure}

\subsection{Model Similarity Analysis}

As shown in Figures~\ref{fig:similar_atlas}-\ref{fig:similar_qwen3}, we evaluate the model similarity between the Agent-Dice fused model, the base model, and models continuously trained on individual datasets. We utilize KL divergence as the metric, where a lower value indicates greater similarity.
The experimental results of the model similarity analysis are presented as heatmaps.

Overall, in the GUI agent domain and the tool-use agent domain, when a base model is trained on more tasks, the KL divergence with the base model increases. This is because the agent continuously learns new knowledge to adapt to new tasks.

In the GUI agent domain, catastrophic forgetting is more pronounced.
So Agent-Dice strikes a balance in parameter shifts across the three datasets. It attains the best performance with only a marginal increase in parameter modification compared to single-dataset training.

In the tool-use agent domain, Agent-Dice exhibits the highest similarity to Qwen3-8B. 
This suggests that Agent-Dice effectively captures the common knowledge intrinsic to the tool-use domain, achieving superior performance with minimal parameter deviation.



\subsection{Overhead Evaluation}

To analyze the additional time overhead introduced by Agent-Dice, 
we conduct an overhead evaluation experiment. 
Specifically, we test different base models in the tool-use agent domain and the GUI agent domain, and report results for scenarios using only the GPU and only the CPU, respectively.
On average, Agent-Dice requires only about one minute of GPU usage or about ten minutes of CPU usage to complete the task.

The statistics are reported in Table~\ref{tab:runtime_comparison}. 
When using a GPU, the overhead of Agent-Dice ranges only from 61.84 to 88.52 seconds. 
Even without a GPU, when using only the CPU, the overhead of Agent-Dice does not exceed one minute. 
Compared to the training process that takes several hours or dozens of hours, the time overhead introduced by Agent-Dice is negligible.
This fully demonstrates that Agent-Dice is a lightweight and efficient agent continual learning solution.

\begin{table}[t]
\centering
\small
\setlength{\tabcolsep}{3pt}
\begin{tabular}{l l c c}
\toprule
\textbf{Model} & \textbf{Domain} & \textbf{GPU Time (s)} & \textbf{CPU Time (s)} \\
\midrule
OS-Atlas-Pro-7B    & GUI & 73.93 & 559.47 \\
Qwen3-VL-8B        & GUI & 83.88 & 463.97 \\
Qwen3-8B           & Tool-use  & 61.84 & 461.03 \\
LLaMA-3.1-8B       & Tool-use  & 88.52 & 1049.82 \\
\midrule
Average   & -  & 77.04 & 633.57 \\
\bottomrule
\end{tabular}
\caption{Computation time comparison across different domains using Agent-Dice with GPU or CPU.} 
\label{tab:runtime_comparison}
\end{table}

\subsection{Comparison With Other Merge Methods}

\begin{table}[t]
\centering
\begin{tabular}{lccc}
\toprule
\textbf{Method} & \textbf{AITZ} & \textbf{AC} & \textbf{GUI-Odyssey} \\
\midrule
Model soups     & 54.10 & 45.65 & 68.40 \\
Adapter-soups   & 57.67 & 50.48 & 70.94 \\
AdapterFusion   & 55.05 & 51.36 & 71.01 \\
Agent-dice      & 57.10 & 51.42 & 72.28 \\
\bottomrule
\end{tabular}
\caption{Performance comparison with other merge methods on GUI agent domain. AC = Androidcontrol.}
\label{tab:model_comparison}
\end{table}

In traditional deep learning, several parameter merging methods have been proposed~\cite{wortsman2022model,chronopoulou2023adaptersoup,pfeiffer2021adapterfusion} to balance optimization objectives in multi-task learning.
In the GUI agent domain, we compare Agent-Dice with these representative approaches.

As shown in Table~\ref{tab:model_comparison}, Agent-Dice consistently outperforms other baselines, as it explicitly identifies shared knowledge while suppressing conflicting updates.
In contrast, Model Soups relies on simple parameter averaging, while Adapter-Soups and AdapterFusion employ low-rank parameter updates, which leads to insufficient learning of new tasks.

\section{Conclusion}
In this work, we identify that the stability–plasticity dilemma in continual learning for LLM-based agents largely arises from the failure to explicitly distinguish between common and conflicting knowledge during the learning process.
To address the challenge, we present Agent-Dice, a novel parameter fusion framework designed to resolve the stability-plasticity dilemma in agent continual learning with minimal computational overhead and parameter updates.
Further, we demonstrate the rationality and effectiveness of Agent-Dice through ablation study, model similarity analysis, and overhead evaluation.
As a lightweight and efficient solution, Agent-Dice paves the way for generalist agents capable of continuous self-iteration in dynamic environments.

\section*{Limitations}
While Agent-Dice has been validated through extensive experiments across different backbone models in the GUI agent domain and the tool-use agent domain, the evaluation centers on a limited set of representative agent scenarios. 
However, this limitation lies in the scope of empirical evaluation rather than in the design of the proposed method itself. 
Future work may explore additional agent domains and task settings to further examine the generality and applicability of Agent-Dice under more diverse and realistic conditions.



\normalem
\bibliography{custom}

\appendix

\section{Proof Datails}
\label{app:proof}
In this section, we provide more detailed proofs for Theorem 1, Theorem 2, and Theorem 3.
\subsection{Detailed Proof of Theorem 1}
Our goal is to prove the update rule $\boldsymbol{\theta}_{\text{new}} = \boldsymbol{\theta}_{\text{pre}} + \sum_{k=1}^K \mathbf{w}_k \odot \boldsymbol{\tau}_k$ approximates a single gradient descent step on a surrogate multi-task objective $\tilde{\mathcal{L}}(\boldsymbol{\theta}) = \sum_{k=1}^K \mathbf{w}_k^\top \mathcal{L}_k(\boldsymbol{\theta})$.

By the linear mode connectivity assumption:
\begin{equation}
\mathcal{L}_k(\boldsymbol{\theta} + \boldsymbol{\delta}) = \mathcal{L}_k(\boldsymbol{\theta}) + \nabla \mathcal{L}_k(\boldsymbol{\theta})^\top \boldsymbol{\delta} + \mathcal{O}(\|\boldsymbol{\delta}\|^2). 
\label{eq:taylor}
\end{equation}

For small \(\|\boldsymbol{\delta}\|\):
\begin{equation}
\mathcal{L}_k(\boldsymbol{\theta} + \boldsymbol{\delta}) \approx \mathcal{L}_k(\boldsymbol{\theta}) + \nabla \mathcal{L}_k(\boldsymbol{\theta})^\top \boldsymbol{\delta}.
\label{eq:linear_approx}
\end{equation}

Each fine-tuning update \(\boldsymbol{\tau}_k\) is obtained via gradient descent with learning rate \(\eta\):
\begin{equation}
\boldsymbol{\tau}_k = -\eta \nabla \mathcal{L}_k(\boldsymbol{\theta}).
\label{eq:update}
\end{equation}

Define scalar weights \(w_k \in \mathbb{R}\) with \(\sum_{k=1}^K w_k = 1\). The fused update is:
\begin{equation}
\boldsymbol{\theta}_{\text{new}} = \boldsymbol{\theta}_{\text{pre}} + \sum_{k=1}^K w_k \boldsymbol{\tau}_k.
\label{eq:fused_update}
\end{equation}

Substituting \eqref{eq:update} into \eqref{eq:fused_update}:
\begin{equation}
\boldsymbol{\theta}_{\text{new}} = \boldsymbol{\theta}_{\text{pre}} - \eta \sum_{k=1}^K w_k \nabla \mathcal{L}_k(\boldsymbol{\theta}_{\text{pre}}).
\label{eq:fused_grad}
\end{equation}

Define the surrogate loss:
\begin{equation}
\tilde{\mathcal{L}}(\boldsymbol{\theta}) = \sum_{k=1}^K w_k \mathcal{L}_k(\boldsymbol{\theta}).
\label{eq:surrogate_loss}
\end{equation}

From \eqref{eq:fused_grad} and \eqref{eq:surrogate_loss}:
\begin{equation}
\boldsymbol{\theta}_{\text{new}} = \boldsymbol{\theta} - \eta \nabla \tilde{\mathcal{L}}(\boldsymbol{\theta}).
\label{eq:gradient_step}
\end{equation}

Thus, the fused update equals one gradient descent step on \(\tilde{\mathcal{L}}\) with step size \(\eta\).

And we define:
\begin{equation}
\boldsymbol{\theta}_{\text{new}} = \boldsymbol{\theta} + \sum_{k=1}^K \mathbf{w}_k \odot \boldsymbol{\tau}_k.
\label{eq:elementwise_update}
\end{equation}

From \eqref{eq:update} and \eqref{eq:elementwise_update}:
\begin{equation}
\boldsymbol{\theta}_{\text{new}} = \boldsymbol{\theta} - \eta \sum_{k=1}^K \mathbf{w}_k \odot \nabla \mathcal{L}_k(\boldsymbol{\theta}).
\label{eq:elementwise_grad}
\end{equation}

From \eqref{eq:gradient_step} and \eqref{eq:elementwise_grad}:
\begin{equation}
    \tilde{\mathcal{L}}(\boldsymbol{\theta}) = \sum_{k=1}^K \mathbf{w}_k^\top \mathcal{L}_k(\boldsymbol{\theta}).
\end{equation}

\subsection{Detailed Proof of Theorem 2}

Our goal is to show that consensus-based filtering yields an update direction whose error probability decays exponentially with the number of consistent tasks, and improves over standard averaging.

Consider a fixed parameter dimension $j$. Let $s^*_j \in \{-1, +1\}$ denote the true descent direction. For each task $k$, define the signed update
\begin{equation}
s_{k,j} = \operatorname{sgn}(\tau_{k,j}).
\end{equation}

Assume
\begin{equation}
P(s_{k,j} = s^*_j) = p, \quad p > \frac{1}{2},
\label{eq:sign_model}
\end{equation}
and $\{s_{k,j}\}_{k=1}^K$ are independent.

Let $\mathcal{S}_j \subseteq \{1,\dots,K\}$ denote the consensus set after filtering, and let $m = |\mathcal{S}_j|$. Define the random variable
\begin{equation}
X = \sum_{k \in \mathcal{S}_j} \mathbb{I}[s_{k,j} = s^*_j],
\end{equation}
with expectation
\begin{equation}
\mathbb{E}[X] = mp.
\end{equation}

An update error occurs if the aggregated direction disagrees with $s^*_j$, i.e.,
\begin{equation}
X \le \frac{m}{2}.
\end{equation}

Applying Hoeffding's inequality yields
\begin{equation}
P\!\left(X \le \frac{m}{2}\right)
\le \exp\!\left(-2m(p - 0.5)^2\right).
\label{eq:consensus_bound}
\end{equation}

Thus, the error probability decays exponentially with the consensus size $m$.

For standard averaging, all $K$ tasks are aggregated, including those with $s_{k,j} \neq s^*_j$. This corresponds to an effective success probability $\tilde{p} \le p$, yielding
\begin{equation}
P_{\text{avg}}(\text{error})
\ge \exp\!\left(-2K(\tilde{p} - 0.5)^2\right),
\quad \tilde{p} < p.
\end{equation}

Comparing with \eqref{eq:consensus_bound}, consensus-based filtering achieves a strictly tighter error bound.

\subsection{Detailed Proof of Theorem 3}

Our goal is to derive the optimal scalar weighting $\mathbf{w}_j$ that maximizes entropy under expected saliency and normalization constraints.

For a fixed parameter dimension $j$, let
\[
u_{k,j} = |\tau_{k,j}|
\]
denote the saliency associated with task $k$. We seek a distribution
\begin{equation}
\mathbf{w}_j = \{ w_{k,j} \}_{k \in \mathcal{S}_j}
\end{equation}
that maximizes the entropy
\begin{equation}
H(\mathbf{w}_j) = - \sum_{k \in \mathcal{S}_j} w_{k,j} \log w_{k,j},
\end{equation}
subject to the constraints
\begin{align}
\sum_{k \in \mathcal{S}_j} w_{k,j} u_{k,j} &= C, \label{eq:saliency_constraint} \\
\sum_{k \in \mathcal{S}_j} w_{k,j} &= 1. \label{eq:normalization_constraint}
\end{align}

We form the Lagrangian
\begin{equation}
\begin{aligned}
\mathcal{L}
= {} & - \sum_{k} w_{k,j} \log w_{k,j} \\
& + \lambda \!\left( \sum_{k} w_{k,j} u_{k,j} - C \right) \\
& + \gamma \!\left( \sum_{k} w_{k,j} - 1 \right).
\end{aligned}
\end{equation}

Taking the partial derivative with respect to $w_{k,j}$ and setting it to zero:
\begin{equation}
\frac{\partial \mathcal{L}}{\partial w_{k,j}}
= - \log w_{k,j} - 1 + \lambda u_{k,j} + \gamma
= 0.
\end{equation}

Solving for $w_{k,j}$ yields
\begin{equation}
\log w_{k,j} = \lambda u_{k,j} + \gamma - 1,
\end{equation}
or equivalently,
\begin{equation}
w_{k,j} \propto \exp(\lambda u_{k,j}).
\end{equation}

Enforcing the normalization constraint \eqref{eq:normalization_constraint}, we obtain
\begin{equation}
w_{k,j}
= \frac{\exp(\lambda u_{k,j})}
{\sum_{m \in \mathcal{S}_j} \exp(\lambda u_{m,j})}.
\end{equation}

By defining $\beta = \lambda$, the optimal solution corresponds to the Boltzmann (Softmax) distribution, completing the proof.

\section{Experimental Details}
\label{app:experiment}
In this section, we provide a comprehensive description of the experimental setup used to evaluate Agent-Dice.
We detail the implementation configurations, action and output formats, and model selections across two representative domains: the GUI agent domain and the tool-use agent domain.
All experiments are conducted under consistent training and evaluation protocols to ensure fair and reproducible comparisons.
\begin{table*}[ht]
    \centering
    \small
    \begin{tabular}{ccc}
     \toprule
    \textbf{Action Type} & \textbf{Action Description} & \textbf{Action Format} \\ 
    \midrule 
    
    CLICK & Click at the specified position. & CLICK \textless point\textgreater[[x-axis, y-axis]]\textless/point\textgreater \\
    TYPE & Enter specified text at the designated location. & TYPE [input text] \\
    SCROLL & Scroll in the specified direction. & SCROLL [UP/DOWN/LEFT/RIGHT] \\
    PRESS\_BACK & Press a back button to navigate to the previous screen. & PRESS\_BACK \\
    PRESS\_HOME & Press a home button to navigate to the home page. & PRESS\_HOME \\
    ENTER & Press the enter button. & ENTER \\
    OPEN\_APP & Open the specified application. & OPEN\_APP [app\_name] \\
    WAIT & Wait for the screen to load. & WAIT \\
    LONG\_PRESS & Long press at the specified position. & LONG\_PRESS \textless point\textgreater[[x-axis, y-axis]]\textless/point\textgreater \\
    COMPELTE & Indicate the task is finished. & COMPELTE \\
    IMPOSSIBLE & Indicate the task is impossible. & IMPOSSIBLE \\
    \bottomrule
    \end{tabular}
    \caption{Action space in our GUI agent domain experiment.}
    \label{action_space}
\end{table*}

\subsection{Output Format}
For the GUI agent domain, we follow the common action space used by existing GUI agents, as shown in Table~\ref{action_space}. 
During evaluation, we adhere to the assessment methods of existing works: for actions with coordinates such as CLICK and LONG\_PRESS, a relative error of less than 14\% is considered correct. For TYPE actions, an F1 score greater than 0.5 is required to be counted as correct. In all other cases, exact matching is necessary for correctness.
And TSR for a task will be 1 only if SR for every single frame within that task is 1.

For the tool-use agent domain, we specify the selectable tools and parameter descriptions in the input prompt. 
The agent directly outputs a list of tool calls, where each element includes the function name along with the corresponding parameter names and values.

\subsection{Model details}
Our experiments are conducted in two domains: the GUI agent domain and the tool-use agent domain.
In the GUI agent domain, we consider both models specialized for GUI manipulation and general-purpose models equipped with GUI interaction capabilities.
To ensure broad and representative evaluation, we select one example from each category, namely OS-Atlas-Pro-7B and Qwen3-VL-8B, for our experiments.

In the tool-use agent domain, not all models are able to follow prompts to invoke tools under a zero-shot setting.
Therefore, we choose Qwen3-8B and Llama-3.1-8B, which demonstrate zero-shot tool-use capability, for evaluation.

\section{How to partition the ToolACE dataset}
\label{app:toolace}

To ensure a balanced distribution of tool capabilities across different partitions, we employ the splitting strategy outlined in Algorithm~\ref{alg:tool_aware_split}. 

In the assignment phase, the ToolACE dataset is first shuffled to eliminate distributional bias. 
We then adopt a greedy allocation approach where each sample is assigned to the subset that minimizes a joint objective: the increment of unseen tools (to promote tool concentration) and the current subset size (to ensure load balancing). 
This tool-aware mechanism effectively prevents the fragmentation of tool occurrences, allowing each subset to specialize in specific functional domains while maintaining uniform data volume.
Subsequently, we execute a density-based intra-subset split to construct robust training and evaluation sets. Within each assigned subset, samples are sorted by tool usage density in descending order; the top portion (determined by ratio $r$) is selected for training to maximize the model's exposure to complex tool-use scenarios. 
For the test set, we identify the tool coverage of the training partition and organize the remaining samples based on their tool novelty relative to the training data. 

\begin{algorithm}[t]
\small
\SetAlgoLined 
\LinesNumbered
\DontPrintSemicolon 
\SetKwInOut{Input}{\textsc{Input}}
\SetKwInOut{Output}{\textsc{Output}}
\SetKwFunction{SortDesc}{\textsc{SortDesc}}
\SetKwFunction{SortAsc}{\textsc{SortAsc}}
\SetKwFunction{Shuffle}{\textsc{Shuffle}}

\caption{ToolACE Split Strategy}
\label{alg:tool_aware_split}

\Input{Dataset $\mathcal{D}$, number of subsets $M$, training ratio $r$}
\Output{Subsets $\{\mathcal{D}_m^{\text{train}}, \mathcal{D}_m^{\text{test}}\}_{m=1}^{M}$}

\BlankLine
\textbf{Phase 1: Tool-Aware Assignment}\;
Initialize $\mathcal{D}_m \leftarrow \emptyset, \mathcal{T}_m \leftarrow \emptyset$ for $m \in \{1, \dots, M\}$\;
\ForEach{$x \in \Shuffle(\mathcal{D})$}{
    $\mathcal{T}(x) \leftarrow \text{extract tools from } x$\;
    $m^* \leftarrow \operatorname*{arg\,min}_{m} \left( |\mathcal{T}(x) \setminus \mathcal{T}_m|,\; |\mathcal{D}_m| \right)$\;
    $\mathcal{D}_{m^*} \leftarrow \mathcal{D}_{m^*} \cup \{x\}; \quad \mathcal{T}_{m^*} \leftarrow \mathcal{T}_{m^*} \cup \mathcal{T}(x)$\;
}

\BlankLine
\textbf{Phase 2: Intra-Subset Splitting}\;
\For{$m = 1$ \KwTo $M$}{
    $\mathcal{D}_m \leftarrow \SortDesc(\mathcal{D}_m, \text{key}=|\mathcal{T}(x)|)$\;
    $N_{\text{train}} \leftarrow \lfloor r \cdot |\mathcal{D}_m| \rfloor$\;
    $\mathcal{D}_m^{\text{train}} \leftarrow \mathcal{D}_m[1 : N_{\text{train}}]$\;
    
    $\mathcal{T}_m^{\text{train}} \leftarrow \bigcup_{x \in \mathcal{D}_m^{\text{train}}} \mathcal{T}(x)$\;
    $\mathcal{D}_{\text{rem}} \leftarrow \mathcal{D}_m \setminus \mathcal{D}_m^{\text{train}}$\;
    $\mathcal{D}_m^{\text{test}} \leftarrow \SortAsc(\mathcal{D}_{\text{rem}}, \text{key}=|\mathcal{T}(x) \setminus \mathcal{T}_m^{\text{train}}|)$\;
}

\Return $\{\mathcal{D}_m^{\text{train}}, \mathcal{D}_m^{\text{test}}\}_{m=1}^{M}$\;

\end{algorithm}

As evidenced in Table~\ref{tab:tool_overlap}, the diagonal entries exhibit significantly higher tool overlap ($25.8\%\sim31.8\%$) compared to the minimal cross-subset leakage ($<3.2\%$) in off-diagonal entries. This distinct boundary confirms that our splitting strategy effectively localizes tool usage within each partition, ensuring that each subset specializes in a distinct functional domain while maintaining strong consistency between its training and testing distributions.

\begin{table}[t]
\centering
\small
\setlength{\tabcolsep}{1.7pt}
\renewcommand{\arraystretch}{1.15}

\begin{tabular}{lcccc}
\toprule
& $\boldsymbol{\mathcal{D}_0^{\text{test}}}$ 
& $\boldsymbol{\mathcal{D}_1^{\text{test}}}$ 
& $\boldsymbol{\mathcal{D}_2^{\text{test}}}$ 
& $\boldsymbol{\mathcal{D}_3^{\text{test}}}$ \\
\midrule
$\boldsymbol{\mathcal{D}_0^{\text{train}}}$ 
& \textbf{103 (29.6 \%)} & 4 (1.2 \%) & 7 (2.1 \%) & 6 (1.8 \%) \\
$\boldsymbol{\mathcal{D}_1^{\text{train}}}$ 
& 8 (2.3 \%) & \textbf{94 (29.1 \%)} & 5 (1.5 \%) & 6 (1.8 \%) \\
$\boldsymbol{\mathcal{D}_2^{\text{train}}}$ 
& 11 (3.2 \%) & 2 (0.6 \%) & \textbf{106 (31.8 \%)} & 5 (1.5 \%) \\
$\boldsymbol{\mathcal{D}_3^{\text{train}}}$ 
& 6 (1.7 \%) & 4 (1.2 \%) & 4 (1.2 \%) & \textbf{84 (25.8 \%)} \\
\bottomrule
\end{tabular}

\caption{Tool overlap statistics between training and testing subsets. 
Diagonal entries (in bold) indicate intra-subset overlap, while off-diagonal entries represent cross-subset overlap.}
\label{tab:tool_overlap}
\end{table}

\section{Case Study}
In this section, we provide examples to show how GUI agents and the tool-use agent work.

\subsection{GUI Agent Case Study}
\begin{figure*}
    \centering
    \includegraphics[width=\linewidth]{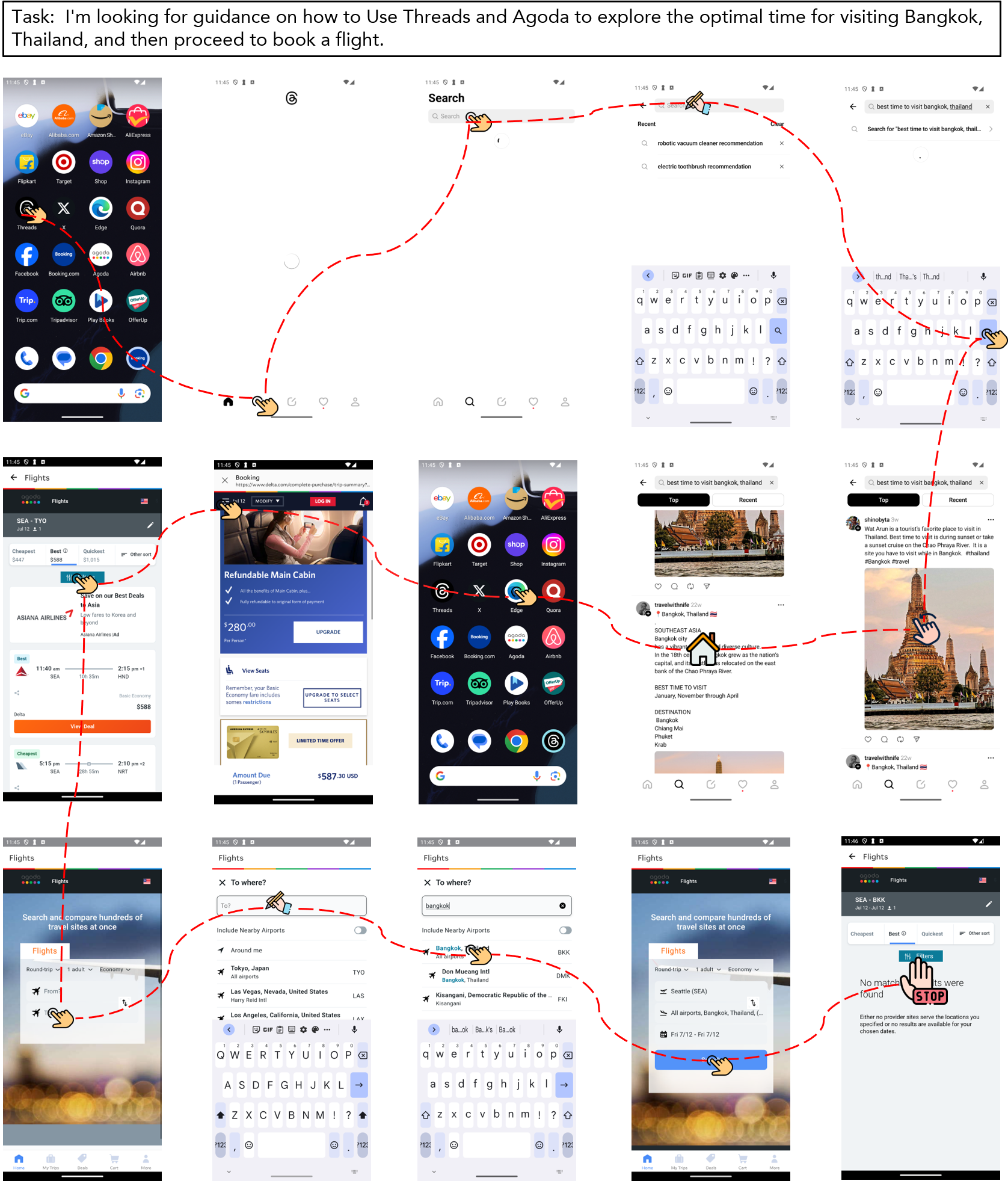}
    \caption{Schematic diagram of the GUI agent case study. The user provides an instruction to the agent, and the agent automatically executes it on a mobile phone or computer by simulating human operations.}
    \label{case_gui}
\end{figure*}

The goal of a GUI agent is to automatically execute instructions on smart terminals by simulating human operations, following user-given commands. 
As shown in Figure~\ref{case_gui}, a user needs guidance for a trip to Bangkok, Thailand, and then requires a flight ticket. The GUI agent first searches for guidance. It clicks the Threads APP icon on the main interface, enters the app, clicks the search button, and inputs text to search for relevant content. 
After finding relevant guidance, it selects the option to return to the main interface using the home button and proceeds to search for flight tickets. Finally, the task is terminated because no matching flight tickets are found.

\subsection{Tool-Use Agent Case Study}
\begin{figure*}
    \centering
    \includegraphics[width=\linewidth]{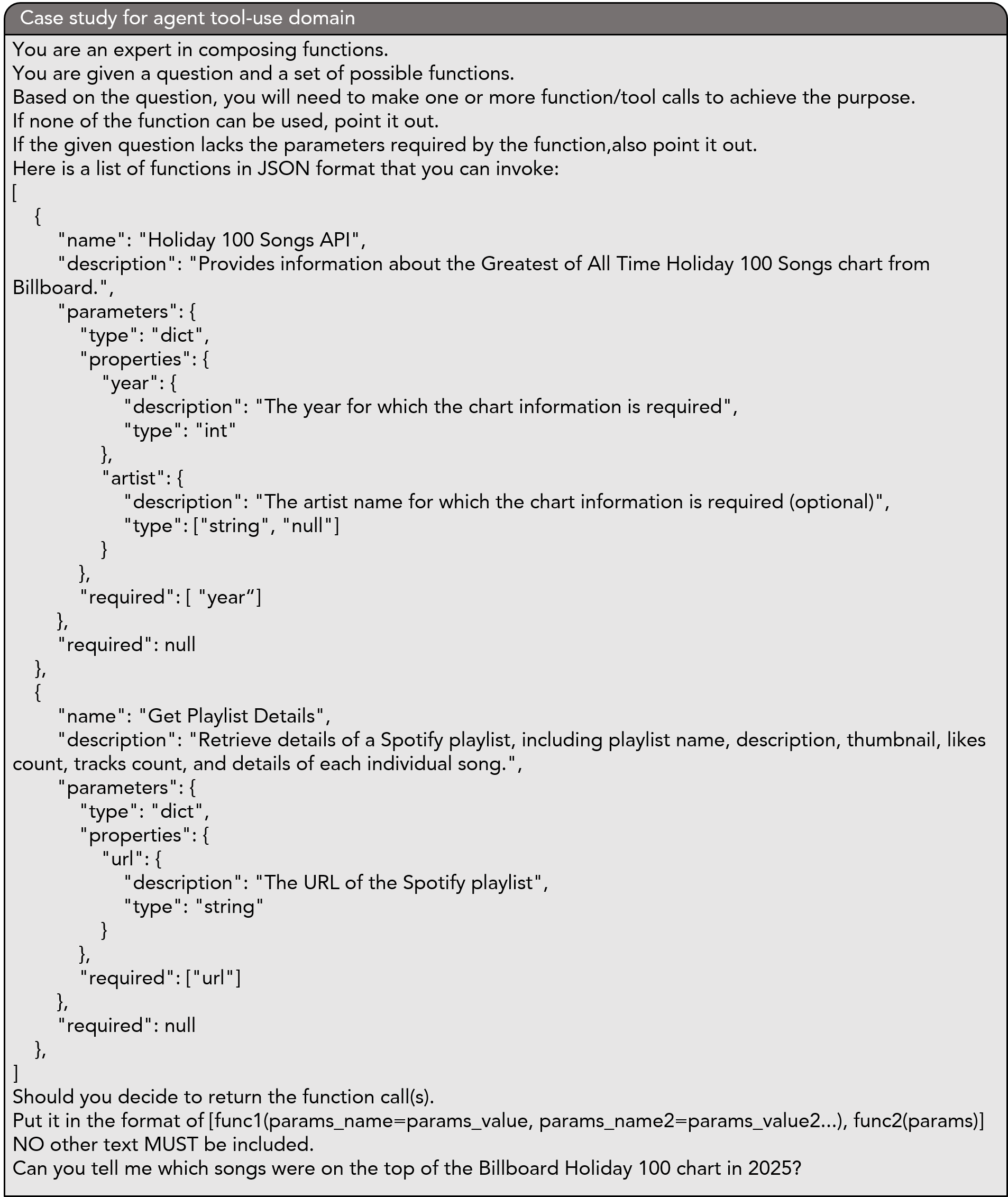}
    \caption{Schematic diagram of tool-use agent. The user provides an instruction that requires the agent to complete the task by calling other tools, demanding the agent to correctly output the tool name, parameter names, and parameter values.}
    \label{case_tool}
\end{figure*}

The goal of tool-use agent is to enable the agent to automatically determine whether it needs to call a tool based on user-given instructions, decide which tool to call, and output the correct function name, parameter names, and parameter values. 
As shown in Figure~\ref{case_tool}, in this prompt, the user asks the agent to inform them which songs were at the top of the Billboard Holiday 100 chart in 2025. 
This falls outside the agent's intrinsic knowledge, so it needs to call a tool. 
Among the tools provided by the user, the Holiday 100 Songs API is the most suitable for the user's instruction. 
Therefore, this function should be called. The Holiday 100 Songs API has two parameters: \texttt{year} and \texttt{artist}. 
The \texttt{artist} parameter is irrelevant to the user's query, so the agent's final response is \texttt{[Holiday 100 Songs API(year=2025)]}.

\end{document}